\documentclass[11pt]{amsart}
\usepackage{amsfonts,amsmath}

\usepackage{xcolor}

\usepackage{tikz}
\usetikzlibrary{shapes.geometric, arrows.meta, positioning, calc}

\tikzstyle{block} = [rectangle, minimum height=1.4cm, minimum width=3.8cm, text centered, draw=black, fill=blue!10, font=\sffamily\small, drop shadow]
\tikzstyle{arrow} = [thick,->,>=Stealth]
\tikzstyle{layer} = [rectangle, rounded corners, minimum height=1.2cm, minimum width=3cm, text centered, draw=black, fill=blue!10]
\tikzstyle{arrow} = [thick,->,>=Stealth]

\usepackage{enumerate,mathrsfs}
\usepackage{amssymb,amsmath,graphicx,amsthm,mathtools,hyperref}

\usepackage{booktabs}

\theoremstyle{plain}
\usepackage{siunitx}

\usepackage{algorithm}
\usepackage{algorithmic,colortbl}

\makeindex
\usepackage[intoc]{nomencl} 

\makenomenclature

\newcommand{\model}{\textsc{NeuroMem-FHP}}

\usepackage{hyperref}
\hypersetup{nesting=true,debug=true,naturalnames=true}
\usepackage{graphicx,amssymb,upref}

\usepackage{subcaption}
\usepackage{enumerate,float}
\usepackage{fullpage}

\begin{document}
\begin{center}
   {\Large \bf \model: A Likelihood-Free Deep Learning Framework for Parameter Estimation of Fractional Hawkes Process} 
\end{center}

\vspace{0.3cm}

\begin{center}
{Neha Gupta}$^{\textrm{a*}}$, {Aditya Maheshwari}$^{\textrm{b}}$

\footnotesize{
\begin{tabular}{l}
$^{\textrm{a}}$ \emph{Operations Management and Quantitative Techniques Area, Indian Institute of Management Indore, Indore 453556, India.}\\
$^{\textrm{*}}$ \emph{Corresponding author: neha.gupta@iimidr.ac.in}
\end{tabular}
}

\vspace{1mm}

\footnotesize{
\begin{tabular}{l}
$^{\textrm{b}}$ \emph{Operations Management and Quantitative Techniques Area, Indian Institute of Management Indore, Indore 453556, India.}\\
\emph{Email: adityam@iimidr.ac.in}
\end{tabular}
}
\end{center}

\email{}
\email{}








			
\begin{abstract}
In this paper, we propose deep learning based \model\ framework for estimating the parameters of the fractional Hawkes process (FHP), a self-exciting point process that captures long-range dependence through a fractional Mittag–Leffler excitation kernel. Two neural architectures, namely a Long Short-Term Memory (LSTM) network and a Transformer, are developed to estimate the model parameters $(\mu,\gamma,\alpha,\beta)$ directly from sequences of inter-arrival times without requiring computationally intensive likelihood optimization. Experiments on synthetic data demonstrate that both neural models significantly outperform the classical Maximum Likelihood Estimation (MLE) method, with the Transformer achieving the highest estimation accuracy (MSE = $0.1634$), followed by the LSTM (MSE = $0.1752$), compared to MLE (MSE = $2.8032$). An ablation study further examines the effects of key hyperparameters on model performance. The proposed framework is also validated on two real-world high-frequency datasets, namely AAPL NBBO transaction data and Montgomery County 911 emergency call records. Using a predictive validation approach, event sequences simulated from the estimated parameters closely reproduce the empirical distribution, tail behavior, and temporal dependence structure of the observed data. These results demonstrate that Transformer-based parameter estimation provides an accurate and efficient alternative to conventional estimation techniques for FHP and offers a promising framework for modeling event-driven systems with long-memory dynamics.

\end{abstract}
\title{}
\maketitle
 \section{Introduction}\label{section:1}
\noindent Counting processes with time-varying intensities have been found very useful in several applications (see \cite{nhpp-appl1,nhpp-appl2,nhpp-appl3}). However, in many real-world applications, the occurrence of future events depends not only on time but also on the history of past events. In particular, the occurrence of the recent event increases the likelihood of observing subsequent events, leading to self-exciting behavior.
To capture such history-dependent dynamics, the Hawkes process was first introduced in the early seventies by Hawkes (see \cite{hawkes1971spectra, hawkes1971point}) and found applications in various fields, for example, in modeling terrorist activities \cite{Porter_2012}, finance (see 
\cite{Bacry_2015,hainaut2017clustered,Hainaut_2019,hautsch2006modelling, bowsher2007modelling,cartea2014buy,chavez2005estimating}), and
seismology (see \cite{hawkes1973cluster,ogata1988statistical,ogata1998space}).
The Hawkes process is a self-exciting counting process with conditional stochastic intensity $\{\Lambda(t|\mathcal{H}_t)\}_{t\geq 0}$, where $\mathcal{H}_t$ represents the history of the counting process and is given by 
\begin{equation*} \label{HP} \Lambda(t|\mathcal{H}_t) = \mu + \alpha \int_0^t f(t-u) \, dN(u), \end{equation*}
\noindent where $\mu >0$ is the baseline intensity, $\alpha>0$ is the jump size, $f(t)$ is the kernel density function of a positive random variable, and $N(t)$ is the counting process. \\
Despite its broad applicability, the classical Hawkes process employs an exponentially decaying self-excitation kernel (see \cite{Bacry_2015}), implying that the influence of past events diminishes rapidly over time and leading to short-range dependence. However, many real-world applications exhibit persistent temporal correlations and long-range dependence that cannot be adequately captured by exponential kernels.
To overcome these limitations, fractional generalizations of Hawkes process have been developed. One formulation of the FHP replaces the exponential kernel with a Mittag-Leffler self-exciting kernel, enabling power-law type memory effects and analytical tractability (see \cite{habyarimana2023fractional}).\\\\
\noindent Another formulation is obtained by subordinating the classical Hawkes process with an inverse stable L\'evy process, leading to a non-Markovian intensity governed by a fractional Fokker–Planck equation (see \cite{hainaut2020fractional}). These fractional versions have applications in modeling seismic aftershock sequences with enhanced clustering behavior (see \cite{davis2024fractional}), as well as in finance, social networks, and other natural phenomena (see \cite{dupret2025fractional}, \cite{habyarimana2023fractional}). For these models to be applied effectively in practice, accurate parameter estimation is essential. Since the model parameters determine the excitation behavior and long-range memory of the process, their reliable estimation is crucial for statistical inference, prediction, and simulation. Existing studies on FHP have focused mainly on approaches based on maximum likelihood estimation (MLE), which often involve complex numerical procedures and high computational cost, especially for large datasets or non-Markovian settings (see \cite{habyarimana2024parameter}). 
These limitations motivate the development of efficient likelihood-free approaches for parameter estimation. \\\\
Recent studies show that deep learning models are effective for parameter estimation in stochastic processes (see \cite{boros2024deep, szarek2021neural, li2023efficient, csanady2024parameter, fein2024comparison, gupta2025neuromemfpp}). The RNN (LSTM) models have been used for parameter estimation in Hawkes process and stochastic differential equations without explicit likelihoods \cite{feng2023deep, lee2023recurrent}. Data-driven CNN–LSTM and recurrent models have shown strong performance for Lévy-driven SDEs \cite{lee2023recurrent}, SDEs with fractional Brownian motion and measurement noise \cite{Wang2003}, SDEs with L\'evy noise \cite{wang2022neural}, and Student L\'evy processes \cite{li2024parameter}. More recently, Transformer-based methods have enabled direct and zero-shot parameter inference in stochastic models using attention mechanisms and in-context learning \cite{Yin2024Transformer, temirkhanov2025context}.\\\\
Although these approaches have demonstrated promising results for a variety of stochastic processes, their application to FHP remains largely unexplored.
Motivated by these advances, we investigate whether deep learning can provide an efficient likelihood-free framework for parameter estimation in the FHP. In the FHP model, the baseline intensity $\mu$, the decay parameter $\gamma$, the branching ratio $\alpha$ and the fractional parameter $\beta$ jointly determine the temporal evolution of the event sequence. In particular, the fractional parameter $\beta\in(0,1)$ governs the strength of long-range dependence induced by the Mittag-Leffler excitation kernel. Since these parameters directly influence the temporal structure of inter-arrival times, sequential neural architectures are well suited for learning the relationship between observed event sequences and the underlying model parameters. 
Deep learning architectures, including LSTM-based models, Transformer-based models, and Physics-Informed Neural Networks (PINNs), have been explored for parameter estimation by combining data-driven learning with structural constraints \cite{liu2023pi,ozalp2023physics,mahar2025attention}. As a type of Recurrent Neural Network (RNN), LSTMs are well suited for sequential data and are effective in capturing long-term dependencies in stochastic processes \cite{hochreiter1997long, greff2016lstm, sagheer2019time, pawar2018stock}. Transformer-based models, which rely on attention mechanisms, have also been successfully applied to mathematical reasoning, symbolic regression, and time-series forecasting \cite{lu2023survey,cranmer2020discovering, lim2021time}, making them promising tools for modeling complex temporal dynamics.\\\\
To the best of our knowledge, this is the first work to investigate deep learning–based parameter estimation for FHP using the proposed \model framework. Unlike existing likelihood-based methods, the proposed framework directly learns the mapping from inter-arrival time sequences to the underlying model parameters without explicit likelihood optimization. The main contributions of this work are summarized as follows:
\begin{itemize}
    \item We propose \model, a deep learning-based framework for parameter estimation of the FHP that employs both RNN (LSTM) and Transformer architectures to estimate model parameters directly from event sequence data, eliminating the need for likelihood-based estimation techniques.
    
    \item Extensive experiments on synthetically generated FHP data demonstrate that the proposed framework achieves a lower mean squared error, improved estimation stability, and faster computation compared to conventional parameter estimation methods.
    
    \item We further validate the proposed \model\ framework on real-world datasets, including emergency call and financial transaction data, demonstrating its ability to accurately capture long-range dependence and time-varying dynamics in event sequences.
\end{itemize}
The remainder of this paper is organized as follows. In Section~\ref{section:2},  we introduce the FHP and present its log-likelihood function. In Section~\ref{section:3}, We describe the proposed deep learning framework, including the RNN (LSTM) and Transformer-based architectures and the synthetic data generation procedure. In Section~\ref{section:4}, we present the experimental results on synthetic datasets, including parameter estimation performance, ablation studies, and computational efficiency comparisons. In Section~\ref{section:5}, we demonstrate the applicability of the proposed framework to real-world emergency call and financial transaction datasets and investigate the effect of window size on parameter estimation and temporal dependence. Finally, in Section~\ref{section:6}, we conclude by summarizing the main findings, discussing the limitations of the proposed approach, and directions for future research.

\section{Background}\label{section:2}
\noindent Here, we define and present some preliminary results and definitions that will be used later in this paper.
\subsection{The Hawkes Process} 
The classical Hawkes process (see \cite{hawkes1971point,hawkes1971spectra}) is a self-exciting counting process whose conditional intensity depends on past events through an exponentially decaying excitation kernel
\begin{equation*}
f(t)=\gamma e^{-\gamma t}, \qquad t>0,
\end{equation*}
where $f(\cdot)$ denotes the excitation kernel. The corresponding conditional intensity is given by
\begin{equation*}
\Lambda(t\mid\mathcal{H}_t)
=
\mu
+
\alpha \sum_{t_i<t}f(t-t_i)
=
\mu
+
\alpha
\sum_{t_i<t}
\gamma e^{-\gamma (t-t_i)},
\qquad t\ge0,
\label{eq:HP}
\end{equation*}
where $\mu>0$ is the baseline intensity, $\alpha\in(0,1)$ is the branching ratio (equivalently, the branching coefficient under this kernel normalization), $\gamma>0$ is the exponential decay parameter, and $\mathcal{H}_t$ denotes the history of the process up to time $t$.

\subsection{Fractional Hawkes Process (FHP)}

The fractional Hawkes process extends the classical Hawkes model by incorporating long-memory effects through a fractional excitation kernel involving the Mittag-Leffler function. The corresponding intensity function (see \cite{hainaut2020fractional}) is given by:
\begin{equation}\label{intensity}
\Lambda(t \mid \mathcal{H}_t)
= \mu + \alpha \sum_{t_i < t} f_{\beta,\gamma}(t - t_i),
\end{equation}
where the excitation kernel is given by
$$
f_{\beta,\gamma}(t)
= \gamma\, t^{\beta - 1} E_{\beta,\beta}\!\left(-\gamma t^{\beta}\right), \quad t>0,
$$
with a fractional parameter $0 < \beta < 1$, which captures the long-range dependence and memory effects in the process. The function $E_{\beta,\beta}(\cdot)$ denotes the Mittag-Leffler function, defined by the series expansion
\begin{equation}
E_{\beta,\beta}(z)
= \sum_{n=0}^{\infty} \frac{z^n}{\Gamma(\beta n + \beta)},
\quad z \in \mathbb{C}, \; \beta > 0.
\end{equation}
\subsection{Likelihood of the Fractional Hawkes Process}

Given a sequence of event times $0 < t_1 < t_2 < \cdots < t_{N(t)}$ observed over
the interval $[0,t]$, the log-likelihood function of the FHP
is given (see \cite{habyarimana2024parameter}) by
\begin{equation*}\label{log_like}
\ell\big(t_1,\ldots,t_{N_t} ; \Theta\big)
= \log\!\left[ L_{t_1,\ldots,t_{N(t)}} \right]
= \sum_{i=1}^{N(t)} \log\!\big(\Lambda(t_i\mid \mathcal{H}_{t_{i}})\big)
- \int_{0}^{t} \Lambda(u\mid \mathcal{H}_{u}) du .
\end{equation*}
Given a set of events \( t_1 < t_2 < \cdots < t_k \), the log-likelihood \( l \) is given by

\begin{equation}
\ell(t_1,\ldots,t_k)
=
\sum_{i=1}^{k}
\log \left[
\mu
+
\alpha \gamma
\sum_{t_j < t_i}
(t_i-t_j)^{\beta-1}
E_{\beta,\beta}\!\left(
-\gamma (t_i-t_j)^\beta
\right)
\right]
-
\mu t_k
-
\alpha
\sum_{i=1}^{k}
\left[
1-
E_{\beta, 1}\!\left(
-\gamma (t_k-t_i)^\beta
\right)
\right].
\end{equation}\\\\
\noindent\textbf{Long-memory and FHP:}
The concept of long memory, also known as long-range dependence (LRD), is well established in the literature with applications in finance, seismology, neuroscience, and related fields (see \cite{karag2004, Ding1993, Pagan1996, DouTaqqu2003, climate2006}). The classical Hawkes process models self-exciting behavior using an exponentially decaying kernel, which leads to short-memory dynamics. To address this limitation, the FHP has been introduced by replacing the exponential kernel with a Mittag-Leffler type fractional kernel, which decays more slowly and induces long-range dependence. This modification preserves the self-exciting structure while introducing non-Markovian behavior and persistent memory effects. The theoretical properties of this process have been studied in \cite{chen2021fractional, habyarimana2023fractional}, where the role of the Mittag-Leffler kernel in generating long-memory behavior is analyzed. Further, \cite{davis2024fractional} provide empirical evidence of long-memory-type behavior in applications such as earthquake modeling.
 \section{Methodology}\label{section:3}
\noindent In this section, we present the proposed \model  frameworks for parameter estimation of the FHP. We first describe the RNN (LSTM) and Transformer architectures and explain how they process sequential inter-arrival time data. We then present the synthetic data generation procedure and introduce the baseline maximum likelihood estimation (MLE) method used for performance comparison. The choice of these architectures is motivated by the temporal dependence and self-exciting nature of the FHP. Unlike the classical Hawkes process, the FHP incorporates fractional memory effects that allow past events to exert a persistent influence on future event occurrences. Consequently, the evolution of the process depends not only on recent events but also on a potentially long history of past arrivals. This combination of self-excitation and long-memory behavior gives rise to complex temporal dependencies that must be captured for accurate parameter estimation.
Before explaining the model architectures, we formulate the parameter estimation problem. Let
$$
\mathbf{T}=(T_1,T_2,\ldots,T_N),
$$
denote a sequence of observed inter-arrival times $T_i$ generated from the FHP, where $N$ is the sequence length. The objective is to learn a nonlinear mapping from the input sequence to the corresponding FHP parameter vector
$$
\boldsymbol{\theta}=(\mu,\gamma,\alpha,\beta).
$$
Specifically, the neural network approximates the function
$$
f_{\omega}:\mathbf{T}\longrightarrow
\hat{\boldsymbol{\theta}}
=
(\hat{\mu},\hat{\gamma},\hat{\alpha},\hat{\beta}),
$$
where $\omega$ denotes the trainable network parameters. The network parameters are optimized by minimizing the mean squared error (MSE) between the predicted and true parameter vectors. The following subsections describe the proposed RNN (LSTM) and Transformer-based parameter estimation architecture in detail.
\subsection{RNN (LSTM) Architecture}
We design an RNN based on the LSTM architecture
 for parameter estimation of the FHP. The choice of the LSTM architecture is motivated by the temporal dependence and long-memory characteristics of the FHP. Unlike the classical Hawkes process, whose excitation decays exponentially, the FHP incorporates a Mittag-Leffler kernel that enables past events to exert a persistent influence on future event occurrences. Consequently, the conditional intensity depends not only on recent events but also on a potentially long history of previous arrivals, resulting in non-Markovian dynamics and long-range temporal dependence. Because of its gated memory structure, the LSTM is able to decide which past information to keep and which to forget over long input sequences, thereby reducing the vanishing-gradient issue commonly seen in standard recurrent neural networks.\\
From a statistical perspective, the fractional parameter $\beta \in (0,1)$ governs the strength of memory in the FHP, with smaller values corresponding to stronger long-range dependence. Since the observed inter-arrival times inherit this temporal dependence, accurate parameter estimation requires a model capable of extracting information distributed across the entire event history rather than relying only on recent observations. The memory cells and gating mechanism of the LSTM enable the network to capture these long-range dependencies, making it particularly suitable for estimating the parameter vector $(\mu,\gamma,\alpha,\beta)$.
\begin{figure}[H]\label{Archi_rnn}
    \centering    

    \includegraphics[width=1\linewidth]{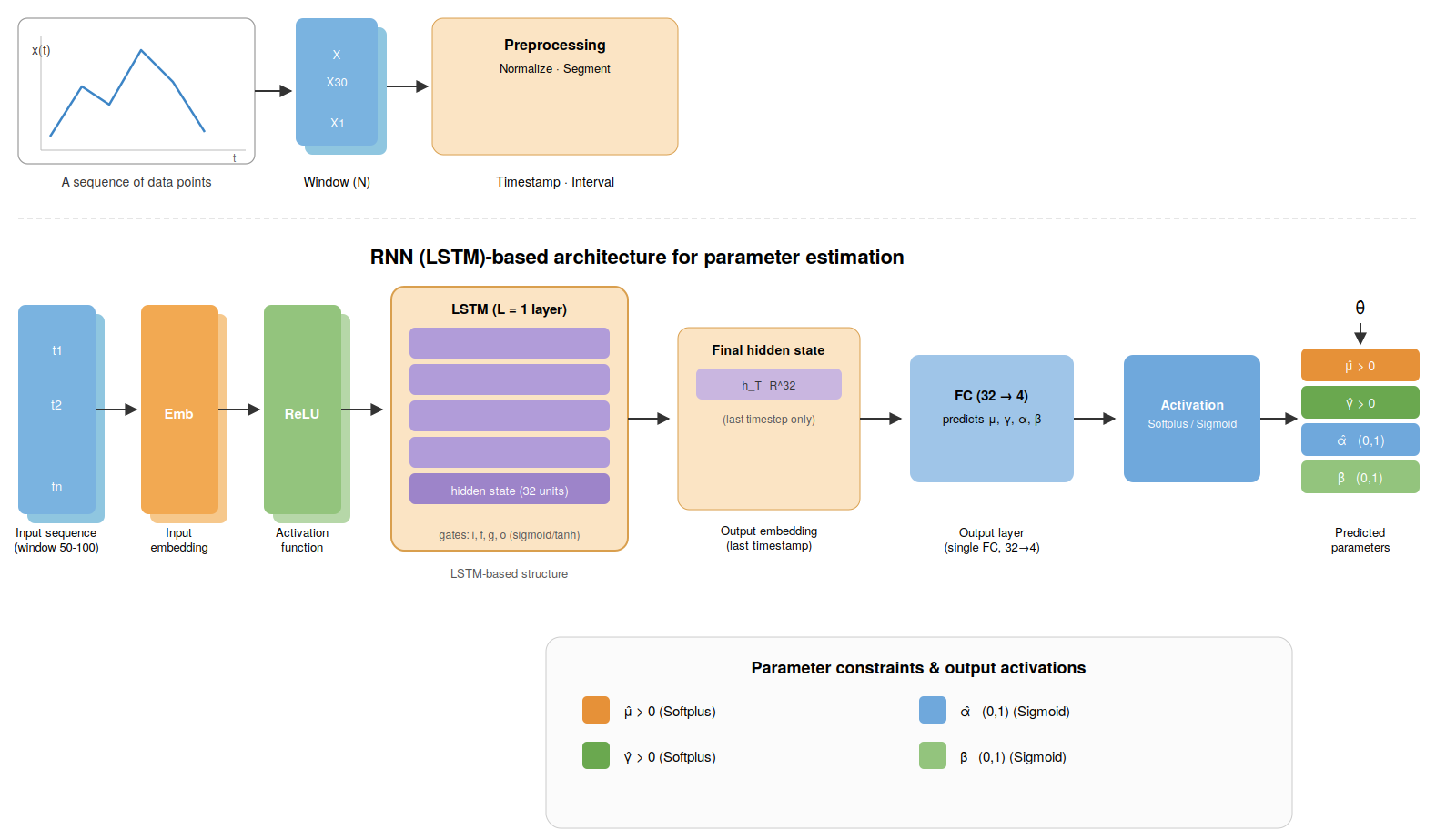}

    \caption{LSTM-based architectures for parameter estimation 
    of the FHP: $\mu > 0, \gamma>0, \alpha \in (0, 1)$ and $\beta \in (0, 1)$.}
    \label{fig:rnn_arch}
\end{figure}

\noindent The proposed architecture is illustrated in Figure~\ref{fig:rnn_arch}. After preprocessing, the event timestamps are converted into inter-arrival times and segmented into fixed-length windows. Each input sequence is then passed through a single LSTM layer consisting of $32$ hidden units (replace with the value used in your implementation). The hidden state generated by the LSTM serves as a compact representation of the temporal dynamics of the observed sequence.
\noindent The LSTM layer computes its outputs using the following set of equations.
\begin{align*}
i_t &= \sigma(W_{ii} x_t + b_{ii} + W_{hi} h_{t-1} + b_{hi}) && \text{(input gate)} \\
f_t &= \sigma(W_{if} x_t + b_{if} + W_{hf} h_{t-1} + b_{hf}) && \text{(forget gate)} \\
g_t &= \tanh(W_{ig} x_t + b_{ig} + W_{hg} h_{t-1} + b_{hg}) && \text{(cell update)} \\
o_t &= \sigma(W_{io} x_t + b_{io} + W_{ho} h_{t-1} + b_{ho}) && \text{(output gate)} \\
c_t &= f_t \odot c_{t-1} + i_t \odot g_t && \text{(cell state update)} \\
h_t &= o_t \odot \tanh(c_t) && \text{(hidden state output)}
\end{align*}

Here, $i_t$, $f_t$, $g_t$, and $o_t$ represent the input, forget, cell, and output gates, respectively, at time $t$. The cell state $c_t$ and hidden state $h_t$ capture the temporal memory of the sequence. The symbol $\odot$ denotes element wise multiplication, and $\sigma$ represents the \texttt{sigmoid} activation function.\\\\
After processing the complete input sequence, the final hidden state of the LSTM is used as a compact representation of the temporal dynamics of the input sequence. This hidden representation of $32$-dimensional is passed directly to a single fully connected layer that predicts the four parameters of the FHP $(\mu, \gamma, \alpha,\beta)$. To satisfy the parameter constraints of the FHP, the \textit{Softplus} activation function is applied to the output corresponding to $(\mu, \gamma)$ to ensure positivity, while the \textit{Sigmoid} activation function, scaled by a factor of $0.9$, is applied to the output corresponding to $\alpha$ and $\beta$, restricting both parameters to the interval $(0,0.9)$. Thus, the network predicts the parameter vector $(\hat{\mu}, \hat{\gamma}, \hat{\alpha},\hat{\beta})$. To satisfy the theoretical constraints of the FHP, the \textit{Softplus} activation function is applied to the baseline intensity $\mu$ and the decay parameter $\gamma$ to guarantee positive predictions, whereas the Sigmoid activation function is applied to the branching ratio $\alpha$ and the fractional parameter $\beta$ to constrain them to the interval $(0,1)$. The network therefore predicts the parameter vector
$
(\hat{\mu},\hat{\gamma},\hat{\alpha},\hat{\beta}).
$
\noindent The proposed LSTM architecture provides sufficient representational capacity for the low-dimensional regression problem considered in this work while remaining computationally efficient. By combining gated memory units with nonlinear feature extraction, the network effectively captures the long-memory and self-exciting characteristics of the FHP, enabling accurate estimation of the underlying model parameters from observed inter-arrival time sequences.\\
In addition, we explain the Transformer-based architecture for estimating the FHP.parameters.

\subsection{Transformer-Based Architecture}

\begin{figure}[H]
\centering 
\includegraphics[width=1\linewidth]{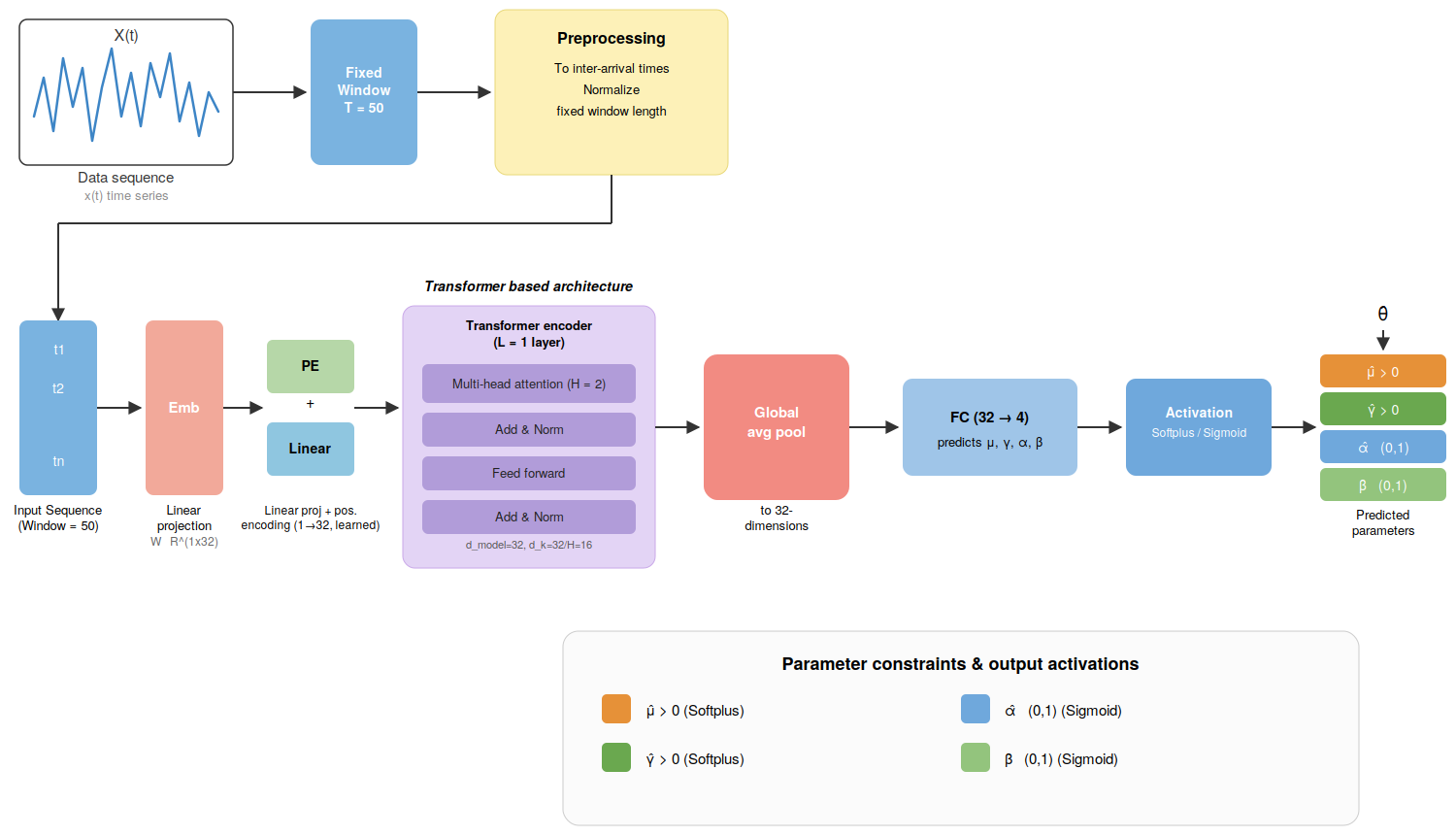} \caption{Transformer-based architectures for parameter estimation of the FHP: $\mu > 0, \gamma>0, \alpha\in (0, 1)$ and $\beta \in (0, 1)$.} \label{fig:all_arch}
\end{figure}

The Transformer-based model~\cite{vaswani2017attention} processes an input
event sequence $\mathbf{T} = [T_1, T_2, \dots, T_n]$ of inter-arrival times.
Unlike recurrent architectures~\cite{hochreiter1997long}, Transformer does not process the sequence step by step. Instead, it uses a self-attention mechanism to learn relationships between all time steps at the same time. This allows parallel computation and helps the model capture direct interactions between different events. The architecture is illustrated in Figure \ref{fig:all_arch}.
\noindent Using the thinning algorithm~\cite{ogata1981}, we generate inter-arrival times
$T_i > 0$ of the FHP and window them into fixed-length input sequences of
shape $(T\times1)$, where $T=50$ denotes the input window length used throughout the experiments.
Each scalar inter-arrival time $T_t$ is first projected into a
$d_{\mathrm{model}} = 32$ dimensional embedding space using a learnable linear projection layer~\cite{nair2010rectified,goodfellow2016deep}:
\begin{equation*}
    W\in\mathbb R^{1\times32}.
\end{equation*}
Since self-attention is permutation-invariant, learnable positional embeddings~\cite{vaswani2017attention} are added to preserve temporal order:
\begin{equation*}
    \tilde{\mathbf{E}}_t \;=\; \mathbf{E}_t + \mathrm{PE}(t),
    \label{eq:pos_enc}
\end{equation*}
where $\mathrm{PE}$ denotes a learnable positional embedding matrix that is optimized jointly with the Transformer parameters during training.

The embedded input sequence is processed by a Transformer encoder consisting of $L = 1$ encoder layer~\cite{vaswani2017attention}. The encoder comprises a multi-head self-attention sublayer followed by a position-wise feed-forward network, with residual connections~\cite{he2016deep} and layer normalization~\cite{ba2016layer} applied after each sublayer. For each of the $H = 2$ attention heads, the embedding dimension is divided equally, resulting in a per-head dimension of
$d_k = d_{\mathrm{model}}/H = 16$. The scaled dot-product attention~\cite{vaswani2017attention} is computed as
$$
    \mathrm{Attention}_h(\tilde{X})
    \;=\;
    \mathrm{softmax}\!\left(\frac{Q_h K_h^{\top}}{\sqrt{d_k}}\right) V_h,
    \label{eq:sdp_attn}
$$
where $Q_h, K_h, V_h$ are learned linear projections of $\tilde{X}$.
The outputs of all heads are concatenated and projected:
\begin{equation*}
    \mathrm{MultiHead}(\tilde{X})
    \;=\;
    \mathrm{Concat}(\mathrm{Attention}_1,\mathrm{Attention}_2)\,W^O.
    \label{eq:multihead}
\end{equation*}
Residual connections and layer normalization are then applied after each
sublayer:
\begin{align*}
    x'  &= \mathrm{LayerNorm}(x + \mathrm{MultiHead}(x)),  \label{eq:res1}\\
    x'' &= \mathrm{LayerNorm}(x' + \mathrm{FFN}(x')),      
\end{align*}
where $\mathrm{FFN}(x) = \mathrm{ReLU}(xW_1+b_1)W_2+b_2$ is the
 feed-forward network based on position. 
 After the Transformer encoder layer, the output sequence
$
x'' \in \mathbb{R}^{T \times 32}
$
is aggregated into a single fixed-length vector using global average pooling.
$$
    \mathbf{h}_{\mathrm{final}}
    \;=\;
    \frac{1}{T}\sum_{t=1}^{T} x''_t
    \;\in\; \mathbb{R}^{32}.
    \label{eq:meanpool}
$$
This pooled representation is passed through a fully connected layer, $\mathrm{FC}(32,4)$, which predicts the four FHP parameters $\mu$, $\gamma$, $\alpha$, and $\beta$. The outputs are transformed to satisfy the parameter constraints:
$\hat{\mu}= \mathrm{softplus}(z_1)+0.01 > 0$, $\hat{\gamma}= \mathrm{softplus}(z_1)+0.01 > 0$, $\hat{\alpha}=0.9\,\mathrm{sigmoid}(z_2)\in(0,0.9)$, and $\hat{\beta}=0.9\,\mathrm{sigmoid}(z_3)\in(0,0.9).$
Here, $\mathrm{softplus}(z)$ ensures strict positivity of $\mu$, while the scaled sigmoid activation constrains $\alpha$ and $\beta$ to the interval $(0,0.9)$, preventing the estimates from approaching the unstable boundary value 1 and maintaining valid ranges of FHP parameters~\cite{kerss2019fractional}.\\

\noindent Transformer-based architectures are particularly suitable for parameter estimation in the FHP because the self-attention mechanism enables each event within the observation window to directly attend to every other event, regardless of their temporal separation. This global receptive field facilitates the learning of long-range temporal dependencies induced by the self-exciting dynamics of the FHP. Consequently, the model effectively captures both local interactions and persistent historical influences, resulting in an accurate and computationally efficient estimation of the process parameters.

\subsection{Data Generation}
To generate training data, we simulate sample paths of the FHP using the thinning algorithm (see \cite{ogata1981}). The algorithm generates a sequence of event times
$0 < T_1 < T_2 < \cdots < T_N \leq T$ from the FHP intensity process given in \eqref{intensity}.
The thinning algorithm proceeds as follows.
At each iteration, an upper bound $M \geq \Lambda(t \mid \mathcal{H}_t)$ on the intensity is
computed as
\begin{equation}
    M
    \;=\;
    \mu + \alpha \sum_{t_i < t+\varepsilon} f_{\beta, \gamma}(t + \varepsilon - t_i),
    \label{eq:upper_bound}
\end{equation}
for a small $\varepsilon > 0$. A candidate event time is then proposed by
drawing a waiting time $E \sim \mathrm{Exp}(1/M)$ and setting
$\tau = t + E$. The candidate is accepted as a true event with probability
\begin{equation}
    p(\tau)
    \;=\;
    \frac{\Lambda(\tau \mid \mathcal{H}_{\tau})}{M}
    \;=\;
    \frac{\mu + \alpha \displaystyle\sum_{t_i < \tau} f_{\beta, \gamma}(\tau - t_i)}{M},
    \label{eq:accept_prob}
\end{equation}
by drawing $U \sim \mathcal{U}(0,1)$ and accepting if $U < p(\tau)$;
otherwise, the candidate is rejected and the procedure repeats from the
updated time $t = \tau$. This continues until $t$ exceeds the horizon $T$.
From the resulting event times $\{t_i\}_{i=1}^{N}$, the inter-arrival times
are computed as
\begin{equation}
    T_i \;=\; t_i - t_{i-1}, \qquad i = 1, 2, \dots, N,
    \label{eq:inter_arrival}
\end{equation}
with $t_0 = 0$. Each sequence $(T_1, T_2, \dots, T_N)$ of
length $T = 100$ forms one training sample, and the corresponding parameter
triple $(\mu, \gamma, \alpha, \beta)$ drawn uniformly from
$\mu \sim \mathcal{U}(0.1, 2.0)$, $\gamma \sim \mathcal{U}(0.1, 2.0)$, $\alpha \sim \mathcal{U}(0.01, 0.9)$, and
$\beta \sim \mathcal{U}(0.01, 0.9)$ constitutes the regression target. A
total of $N_{\mathrm{train}} = 100{,}000$ such pairs are generated and cached
for all subsequent experiments.
In this simulation study, we generate synthetic sample paths of the FHP to construct a training dataset for the LSTM- and Transformer-based parameter estimation models. A large number of trajectories are simulated, with the parameters randomly sampled from their respective admissible ranges to ensure broad coverage of the parameter space. This synthetic dataset is then used to train and evaluate the ability of the proposed models to estimate the underlying FHP parameters $(\mu,\gamma,\alpha,\beta)$ from observed inter-arrival time sequences.
\section{Results and Ablation Study}\label{section:4}
 \noindent In this section, we evaluate the performance of the proposed \model frameworks, namely RNN (LSTM) and Transformer-based parameter estimation frameworks for the FHP using synthetic data. We also present ablation studies and summarize results in tables to compare performance with classical MLE.
\subsection{Results on Synthetic Data}
In this study, we conducted a series of experiments to evaluate the LSTM's  and Transformer performance on synthetic 
data. Our experimental setup, comparison with the baseline MLE estimator, ablation studies, and computational efficiency assessment. To observe the effectiveness of the proposed approaches, we generated a large synthetic dataset by simulating sample paths of FHP over a broad range of parameter values.
For each simulation, the parameters $\mu > 0$, $\gamma > 0$, $\alpha \in (0,1)$, and $\beta \in (0,1)$ were independently sampled from uniform distributions within the ranges $[0.1,2.0]$, $[0.10,5.0]$, $[0.01,0.9]$, and $[0.01,0.9]$, respectively.
Each realization consisted of 100 inter-arrival times, and a total of 100,000 FHP samples were generated. Event-time sequences were obtained via cumulative sums of inter-arrival times, while the inter-arrival times themselves were used as input features. Each sample was paired with its corresponding ground-truth parameter vector $(\mu,\gamma,\alpha,\beta)$ for supervised learning. We consider two deep learning architectures for parameter estimation: RNN (LSTM) and  Transformer-based model.\\\\

\begin{figure}[H]\label{rmse_trans_lstm}
    \centering
    \includegraphics[width=0.8\linewidth]{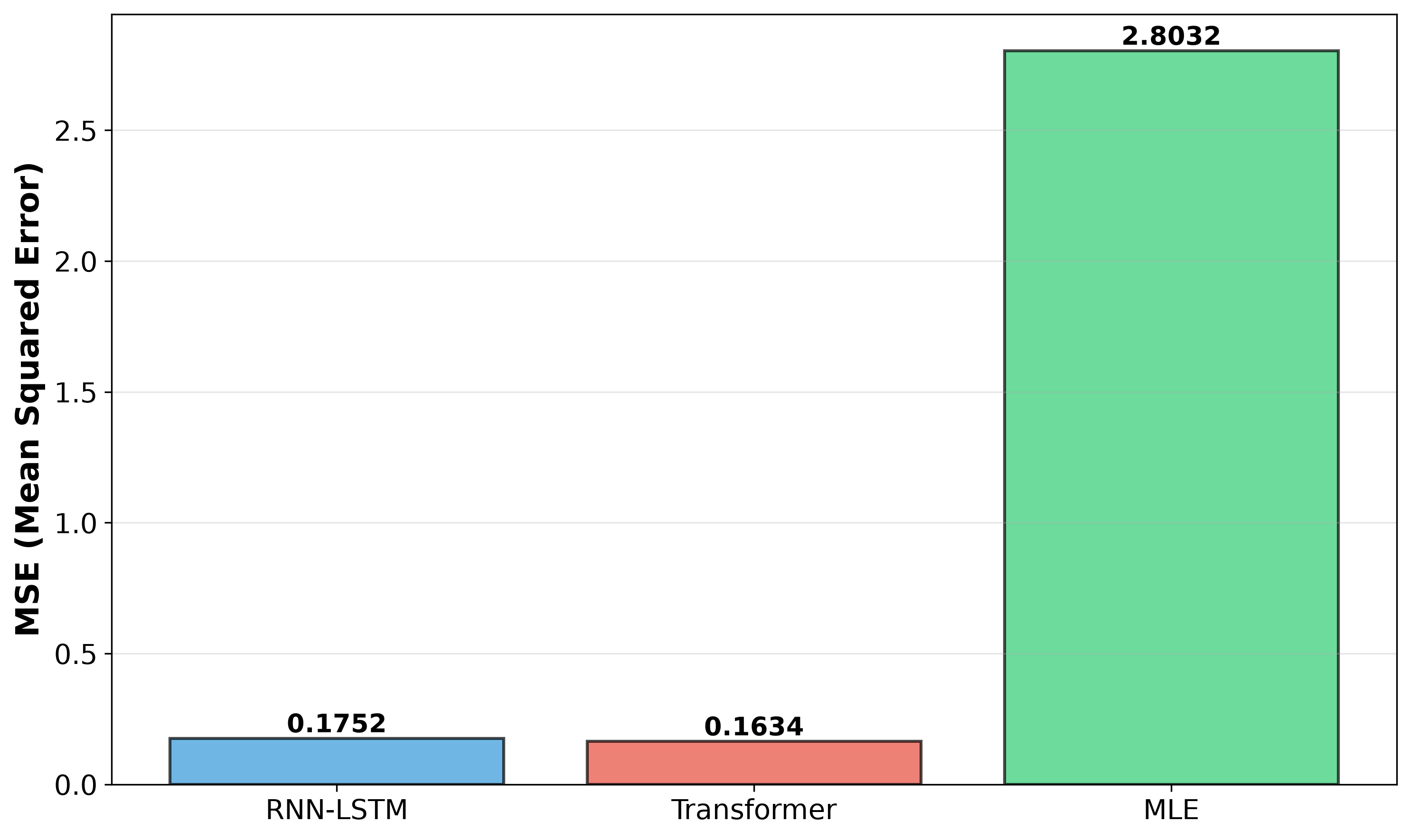}
     \caption{Comparison of parameter estimation accuracy for the FHP using RNN-LSTM, Transformer, and MLE. The deep learning models significantly outperform the classical MLE approach, with the Transformer achieving the lowest Mean Squared Error.}
    \label{over_rmse}
\end{figure}

\noindent The LSTM model consists of a single LSTM layer with $32$ hidden units, followed by a fully connected layer of size $32$ with \textit{ReLU} activation, and a final output layer with four neurons corresponding to $(\hat{\mu}, \hat{\gamma}, \hat{\alpha}, \hat{\beta})$. To enforce parameter constraints, the \textit{Softplus} activation function is applied to $\mu$ and $\gamma$ to ensure positivity, while the \textit{Sigmoid} activation function is applied to $\alpha$ and $\beta$ to restrict them to the interval $(0,1)$.\\
\noindent The Transformer model first projects the input sequence into a $32$-dimensional embedding space and processes it using Transformer encoder layers with self-attention to capture long-range temporal dependencies. This is followed by a Layer Normalization layer, a fully connected size layer $32 \rightarrow 32$ with \textit{ReLU} activation, and a final four-neuron output layer with the same constraint-preserving activations as the LSTM model.
The dataset is randomly split into 80\% training and $20\%$ testing sets. Both models are trained for $100$ epochs using the Adam optimizer \cite{kingma2014adam} with a learning rate of $10^{-3}$ and mean squared error (MSE) loss.\\
\noindent After training, both models are evaluated on the test set and compared with a numerically stabilized MLE, where parameter constraints are enforced and numerical instabilities are handled through clipping and stabilized logarithmic computations. Only convergent and valid estimates are retained for comparison. Figure~\ref{over_rmse} shows that the Transformer achieved the lowest MSE of $0.1634$, followed by the LSTM with an MSE of $0.1752$, while the MLE produced a substantially higher MSE of $2.8032$.




These results indicate that deep learning approaches significantly outperform the classical MLE in estimating the parameters of the FHP. Among the neural models, the Transformer provides the highest estimation accuracy, suggesting that its self-attention mechanism is particularly effective at capturing the long-range dependence induced by the Mittag-Leffler excitation kernel. The superior performance of both neural architectures highlights their ability to learn complex temporal patterns directly from event sequences and accurately recover the underlying FHP parameters.

 \subsection{Ablation Studies}
 Here, we analyzed several results to demonstrate the performance of the Transformer model for parameter estimation of the FHP. These results include the variation of the root mean square error (RMSE), the mean absolute error (MAE), and the coefficient of determination ($R^2$) with respect to the number of training epochs, sample sizes, sequence lengths, batch sizes, and hidden sizes.\\
 To address the heavy-tailed nature of the Mittag--Leffler kernel in the FHP, we modify the preprocessing of the temporal sequence. The event times $T$ are first converted into inter-arrival times $\Delta t$, and a logarithmic transformation $\Delta t_{\text{log}} = \log(1 + \Delta t)$ is applied to reduce the influence of extreme values. The transformed sequence is then standardized before training. This preprocessing step compresses large outliers while preserving the temporal ordering of events, enabling the Transformer to better capture variability in the sequence.

From a statistical perspective, the logarithmic transformation reduces heavy-tailed behavior and stabilizes the variance of the inter-arrival distribution, while standardization improves numerical stability during training. Importantly, these preprocessing steps do not alter the underlying dependence structure of the sequence, allowing the model to learn the long-range temporal patterns characteristic of fractional point processes.\\
\begin{figure}[htbp]
\centering

\begin{subfigure}[b]{0.48\textwidth}
    \centering
    \includegraphics[width=\textwidth]{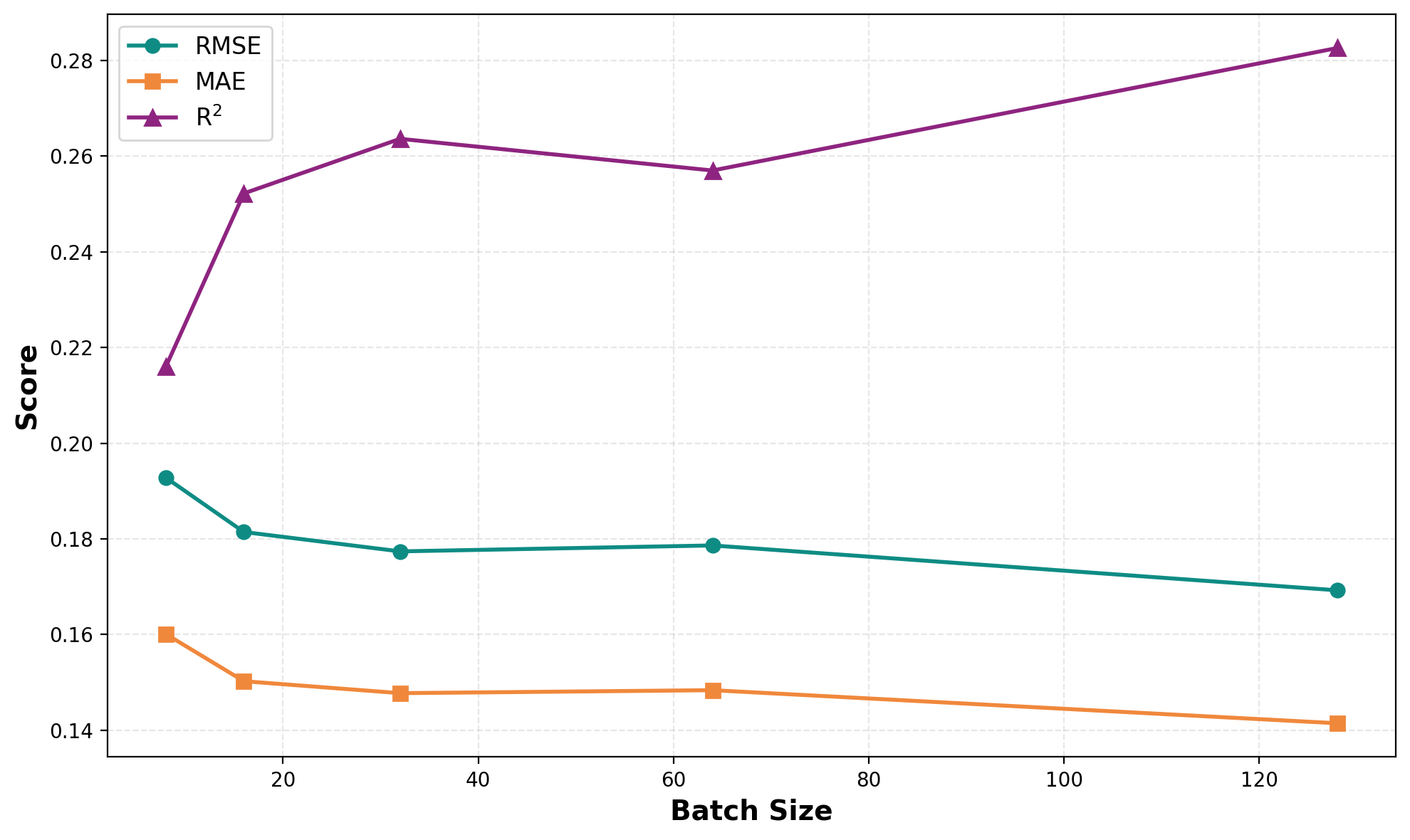}
    \caption{Effect of batch size on model performance.}
    \label{fig:batch_ablation}
\end{subfigure}
\hfill
\begin{subfigure}[b]{0.48\textwidth}
    \centering
    \includegraphics[width=\textwidth]{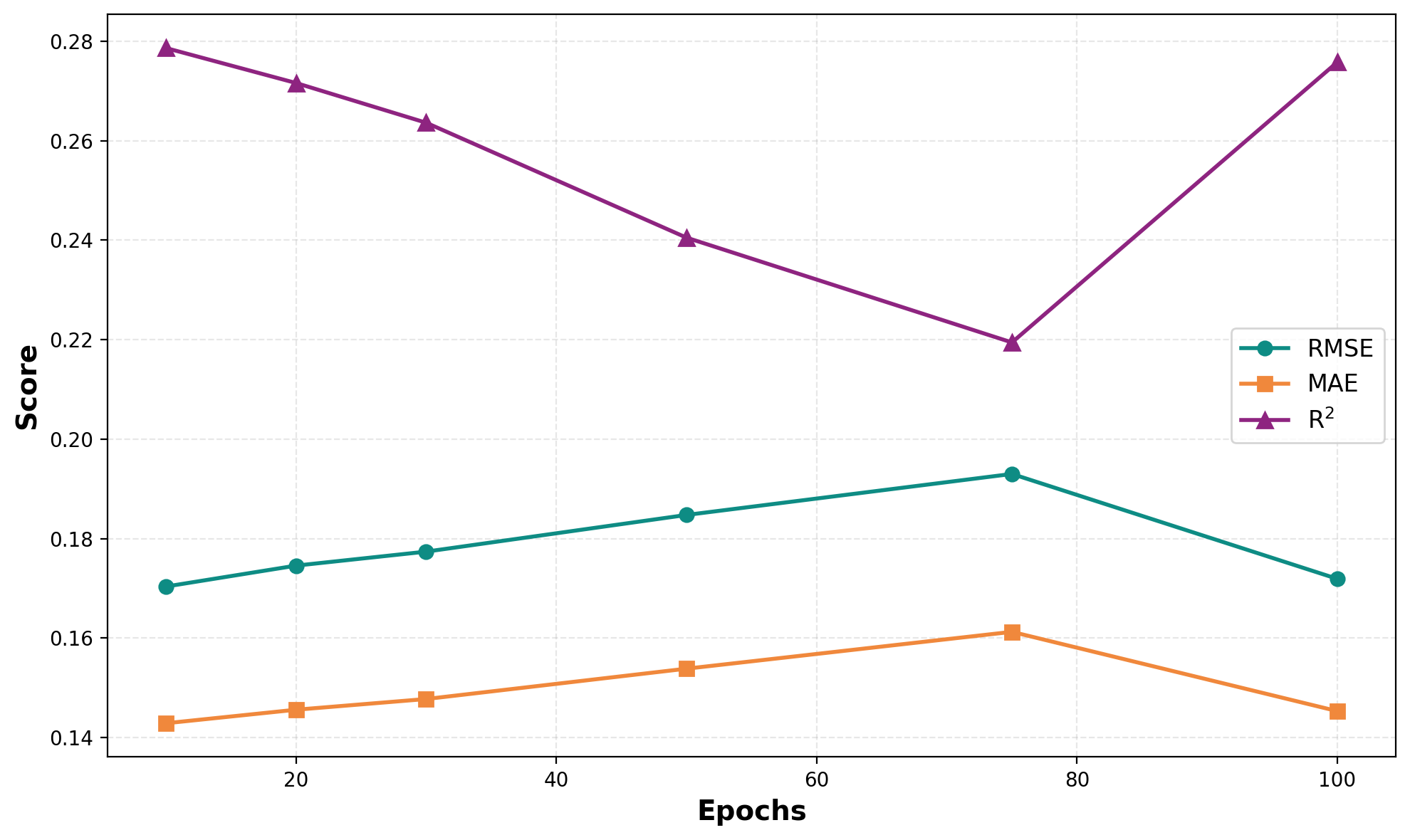}
    \caption{Effect of training epochs on model performance.}
    \label{fig:epoch_ablation}
\end{subfigure}

\vspace{0.3cm}

\begin{subfigure}[b]{0.48\textwidth}
    \centering
    \includegraphics[width=\textwidth]{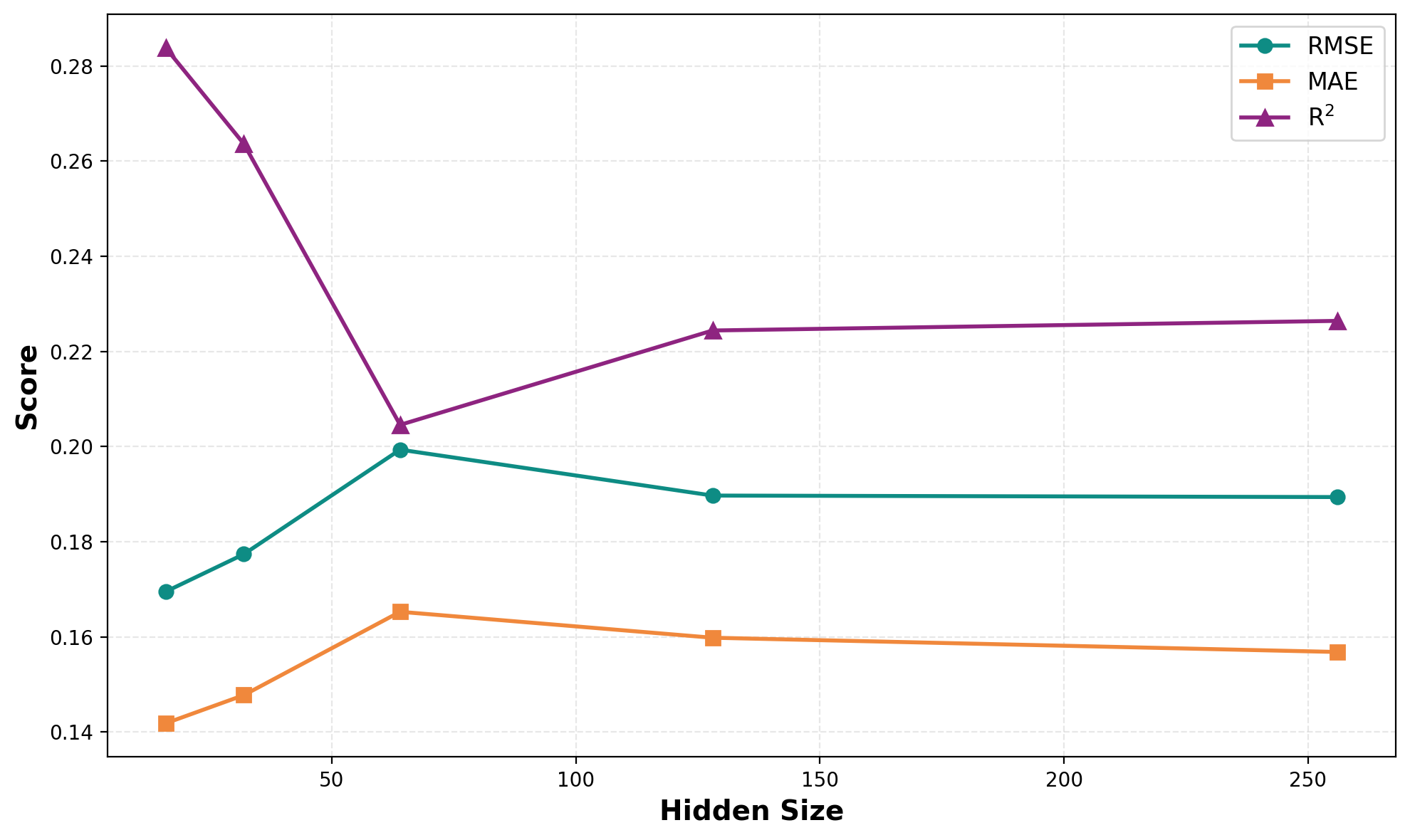}
    \caption{Effect of hidden layer size on model performance.}
    \label{fig:hidden_ablation}
\end{subfigure}
\hfill
\begin{subfigure}[b]{0.48\textwidth}
    \centering
    \includegraphics[width=\textwidth]{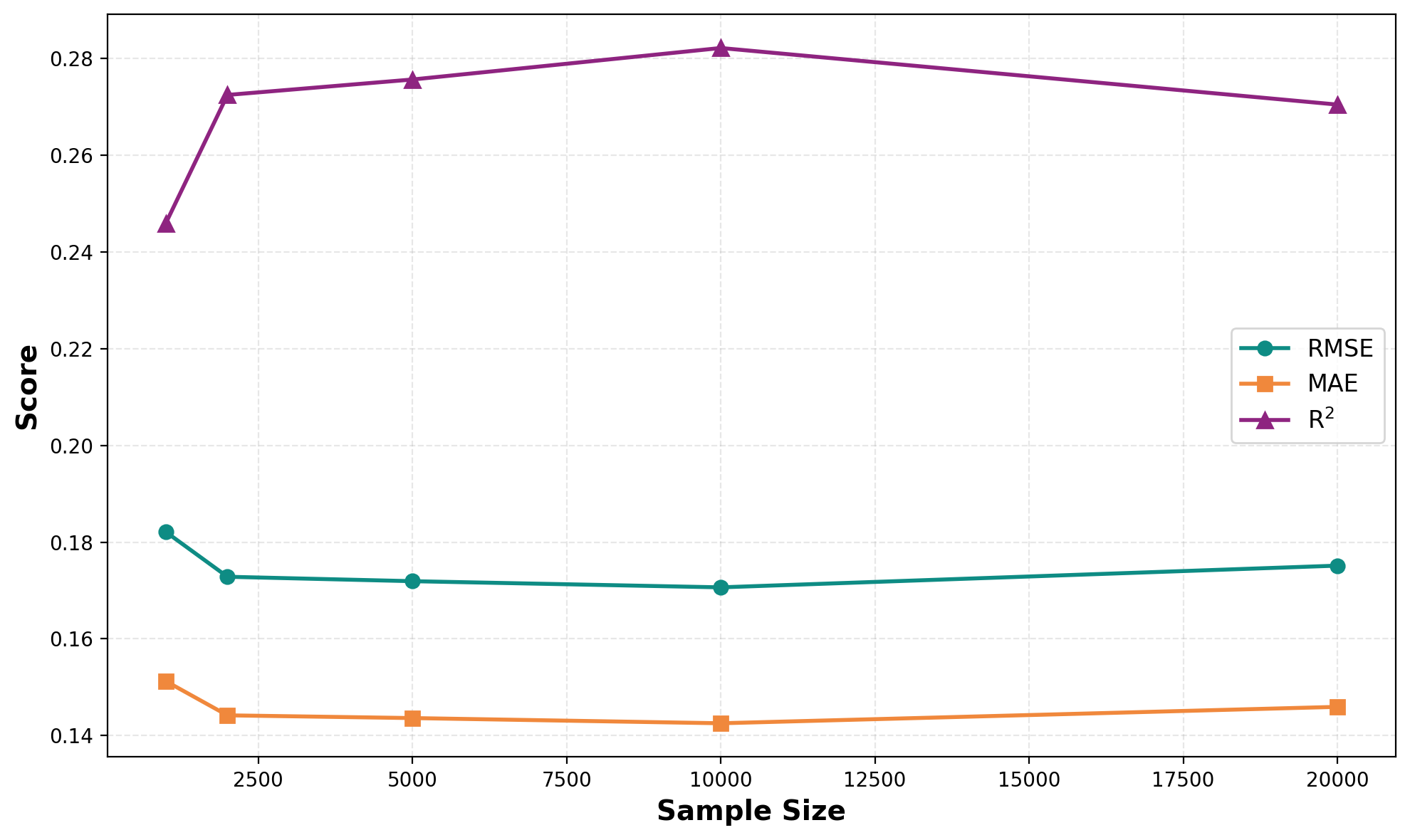}
    \caption{Effect of training sample size on model performance.}
    \label{fig:sample_ablation}
\end{subfigure}

\caption{Ablation study of the Transformer-based FHP model. The plots illustrate the impact of batch size, number of epochs, hidden layer size, and training sample size on score metrics.}
\label{fig:ablation_study}
\end{figure}
 The ablation study presented in Figure~\ref{fig:ablation_study} systematically analyzes the sensitivity of the Transformer-based FHP estimator with respect to key hyperparameters, namely batch size, number of training epochs, hidden layer size, and training sample size. The objective of this analysis is to understand how architectural and optimization choices influence the stability and predictive accuracy of the proposed model.
\noindent The top-left plot in Figure~\ref{fig:ablation_study} illustrates the effect of batch size on model performance. As the batch size increases from $8$ to $128$, we observe a consistent improvement in estimation accuracy. Specifically, the RMSE decreases from approximately $0.193$ to $0.169$, while the MAE reduces from about $0.160$ to $0.141$. In parallel, the $R^2$ score improves from 0.216 to 0.283. This trend indicates that larger batch sizes lead to more stable gradient estimates during training, reducing the stochastic noise in parameter updates.\\
The top-right plot Figure~\ref{fig:ablation_study} shows the impact of the number of training epochs. In the initial phase, increasing epochs from $10$ to $60$ leads to a noticeable reduction in RMSE (from approximately $0.182$ to $0.171$) and a corresponding improvement in MAE and $R^2$ values. This behavior reflects the model’s learning phase, where it captures the underlying temporal structure of the FHP. After around $60–80$ epochs, the performance begins to stabilize, indicating convergence.\\
\noindent The bottom-left plot Figure~\ref{fig:ablation_study} examines the influence of hidden layer size, which controls the model capacity. As the hidden dimension increases from $16$ to $32$, the model initially benefits from an increase in representational power, as reflected by improved values $R^2$, reaching a peak of approximately $0.281$. However, further increases in hidden size do not lead to consistent gains and in some cases slightly degrade performance. This suggests a trade-off between expressiveness and overfitting: while larger hidden layers allow the model to capture more complex temporal dependencies, they may also introduce additional variance, reducing generalization performance on unseen data. Therefore, a moderate hidden size (around $32$ units) appears to provide the best balance.\\
\noindent Finally, the bottom-right plot Figure~\ref{fig:ablation_study}  evaluates the effect of training sample size. Increasing the number of training samples from $1,000$ to $20,000$ consistently improves model performance across all metrics. The RMSE decreases from approximately $0.182$ to $0.175$, MAE improves from $0.151$ to $0.145$, and $R^2$ increases from $0.246$ to $0.271$. This improvement is expected, as larger datasets provide richer representations of the underlying stochastic process, allowing the model to learn more robust and generalized parameter mappings. In particular, the reduction in variance due to increased data size plays a crucial role in stabilizing parameter estimation for fractional point processes.\\
\noindent Overall, the ablation results demonstrate that the Transformer-based estimator is sensitive to training configuration, with performance improvements driven primarily by moderate-to-large batch sizes, sufficient training epochs, balanced hidden capacity, and adequately large datasets. These findings highlight the importance of careful hyperparameter selection when modeling complex temporal processes such as the FHP.\\

\begin{table}[h]
\centering

\begin{tabular}{lcccc}
\hline
\textbf{Method} & \textbf{MSE} & \textbf{RMSE} & \textbf{MAE}  & \textbf{Time (s)} \\
\hline
RNN (LSTM)      & 0.13967 & 0.373725 & 0.299104 & 3.192945 \\
Transformer   & 0.117398 & 0.342633 & 0.274233& 1.877414 \\
MLE           & 2.901081 & 1.703256 & 0.731345 & 0.824175 \\
\hline
\end{tabular}
\caption{Performance comparison of different methods for parameter estimation}
\label{tab:method_comparison}
\end{table}
Furthermore, we investigate the statistical properties of the Transformer-based estimator and evaluate its sampling behavior for four parameters $(\mu,\gamma,\alpha,\beta)$ of the FHP. For this purpose, $100000$ sample paths were generated, each consisting of a sequence of $100$ events, using the true parameter values. The overall estimation performance is summarized in Table~\ref{tab:method_comparison} in terms of MSE, RMSE, MAE, $R^2$, and computational time.\\

\noindent In the Table~\ref{tab:method_comparison}, both deep learning-based methods substantially outperform the classical MLE across all error metrics. In particular, the Transformer achieves the best overall performance, yielding the lowest MSE, RMSE, and MAE, indicating superior accuracy in recovering the full parameter vector $(\mu,\gamma,\alpha,\beta)$. In terms of computational efficiency, the Transformer is more computationally efficient than the RNN (LSTM), while both deep learning-based methods require more computation than the classical MLE approach. The RNN (LSTM) also provides competitive performance and consistently improves upon MLE; however, it is outperformed by the Transformer, highlighting the advantage of self-attention mechanisms in capturing long-range dependencies in FHP dynamics.
 \section{Application on real data}\label{section:5}
 
\noindent In this section, we present the practical applicability of our proposed method to real world high-frequency datasets. 
We focus on evaluating how well the trained Transformer model can capture the temporal dynamics and statistical properties of empirical inter-arrival time series.

 \subsection{Real Data Validation}
 To validate the practical applicability of our method, we applied the trained Transformer 
to two real world high-frequency datasets that exhibit temporal clustering and potential 
long-range dependence. To discuss the empirical results,  we consider two distinct datasets. 
The first dataset corresponds to the NBBO quotes of AAPL, containing high-frequency intraday transaction timestamps for a specific trading day. The timestamps are converted into microseconds, and the differences between consecutive timestamps are used to obtain the inter-arrival times. \\
\noindent The preprocessing of the inter-arrival times is guided by the requirements of the FHP  model. The event timestamps from the NBBO feed are first ordered chronologically, and the corresponding inter-arrival times are computed. To ensure temporal consistency and valid event dynamics, zero and negative time differences are removed. The resulting sequence of positive inter-arrival times is then used as input to the parameter estimation framework.
Inter-arrival times exceeding 100 seconds are further excluded, as such large gaps correspond to intraday market inactivity periods rather 
than genuine tick arrivals; retaining them 
would artificially inflate the empirical 
variance and distort the tail behavior of 
the distribution, thus biasing the 
estimated parameters $(\hat{\mu}, \hat{\gamma}, \hat{\alpha}, \hat{\beta})$ 
away from the true arrival dynamics. A 
$\log_{10}$ transformation is then applied, 
which stabilizes the heavy-tailed structure 
of the inter-arrival times across multiple 
timescales and brings the data into the same 
normalized input space used during synthetic 
training, preserving the validity of the 
zero-shot transfer to real data.\\
\noindent To evaluate the performance of the proposed model on real data, we adopt a predictive validation approach. The trained Transformer model is applied to the inter-arrival time series, where the data are divided into sliding windows and the model estimates the FHP parameters $(\hat{\mu}, \hat{\gamma}, \hat{\alpha}, \hat{\beta})$ for each window. Using these estimated parameters, new event sequences are simulated from the FHP and their statistical properties are compared with those of the real data. 



\begin{figure}
\centering

\begin{subfigure}[b]{0.48\textwidth}
    \centering
    \includegraphics[width=\textwidth]{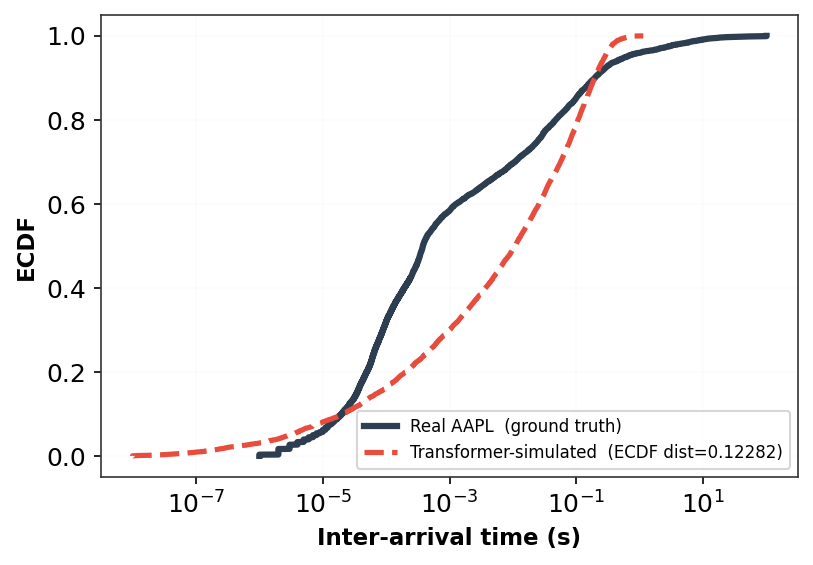}
    \label{fig:loss}
\end{subfigure}
\hfill
\begin{subfigure}[b]{0.48\textwidth}
    \centering
    \includegraphics[width=\textwidth]{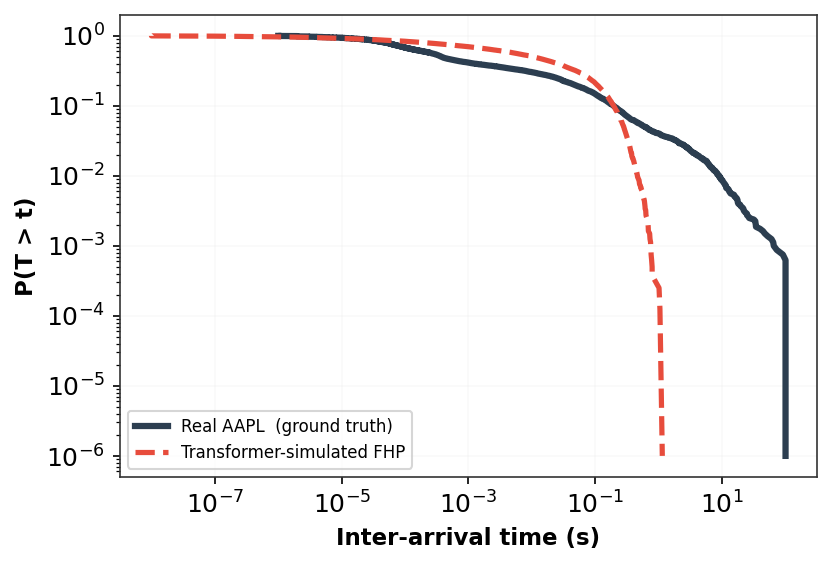}
    \label{fig:boxplot}
\end{subfigure}

\vspace{0.3cm}

\begin{subfigure}[b]{0.48\textwidth}
    \centering
    \includegraphics[width=\textwidth]{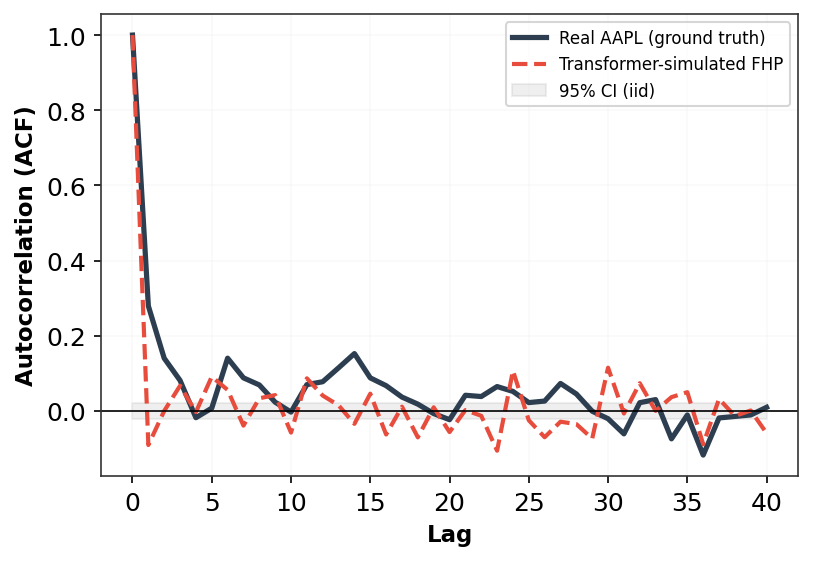}
    \label{fig:beta}
\end{subfigure}
\hfill
\begin{subfigure}[b]{0.48\textwidth}
    \centering
    \includegraphics[width=\textwidth]{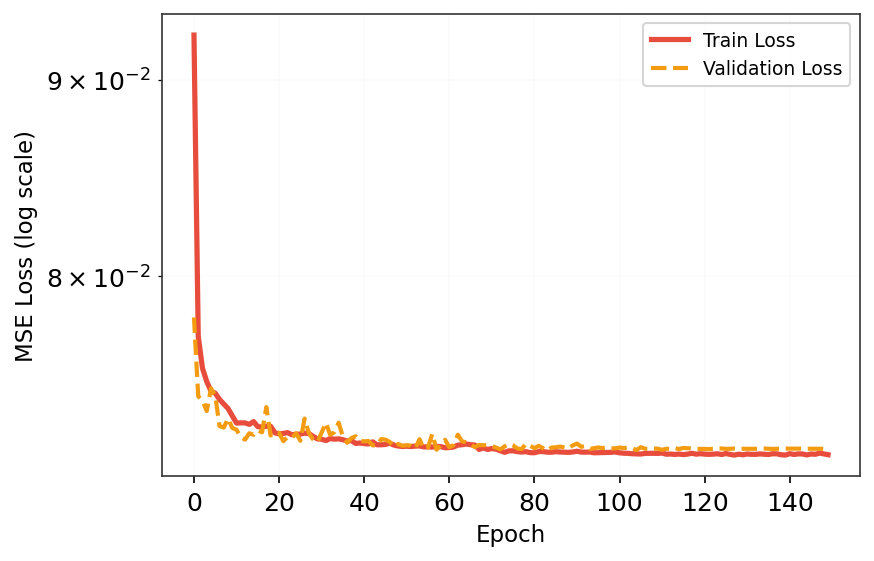}
    \label{fig:ecdf}
\end{subfigure}

\caption{Performance evaluation of the Transformer-based FHP framework: (a) ECDF comparison between real (AAPL) and simulated inter-arrival times. (b) Survival probability comparison demonstrating the ability of the model to reproduce the tail behavior of the empirical distribution. (c) ACF comparison showing that the simulated sequence preserves the dependence structure observed in the real data (AAPL). (d) Training and validation loss curves illustrating stable convergence and good generalization performance.}
\label{NBBO}

\end{figure}

As shown in Figure~\ref{NBBO}, the empirical cumulative distribution function (ECDF) of the simulated data closely matches the distribution of the real inter-arrival times, indicating that the model captures the overall distributional behavior. The survival probability plot further shows that the simulated process reproduces the tail characteristics of the real data. These results indicate that the parameters estimated by the Transformer model are able to reproduce the key statistical and temporal properties of the real event process.




\noindent Next, we analyze the autocorrelation function (ACF) to compare the temporal dependence structure of the real datasets with the sequences simulated using the parameters estimated by the Transformer based FHP model. As shown in the left-bottom panel of Figure \ref{NBBO}, which corresponds to the NBBO quotes of AAPL, the ACF of the real inter-arrival times exhibits a sharp decline immediately after lag $0$ and then fluctuates around zero for higher lags. This pattern indicates a weak short-range dependence, which is commonly observed in high-frequency financial event data. The ACF obtained from the Transformer-simulated FHP closely follows the empirical curve and remains largely within the $95\%$ confidence interval, indicating that the estimated parameters successfully capture the temporal dependence structure of the market data.\\
Similarly, the bottom-right panel of Figure \ref{NBBO} presents the evolution of the training and validation losses. A sharp decline is observed during the early training stages, after which both curves stabilize and converge to nearly identical values. The minimal gap between the two curves suggests that the network has learned a robust representation while maintaining good generalization performance on unseen data.




\hfill




\begin{figure}[htbp]
\centering

\begin{subfigure}[t]{0.48\textwidth}
    \centering
    \includegraphics[width=\linewidth]{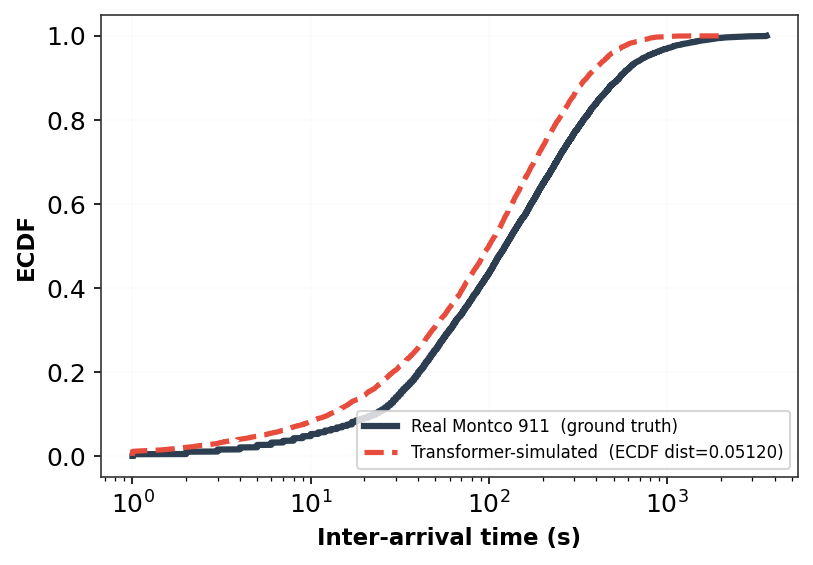}
    \label{fig:911_ecdf}
\end{subfigure}
\hfill
\begin{subfigure}[t]{0.48\textwidth}
    \centering
    \includegraphics[width=\linewidth]{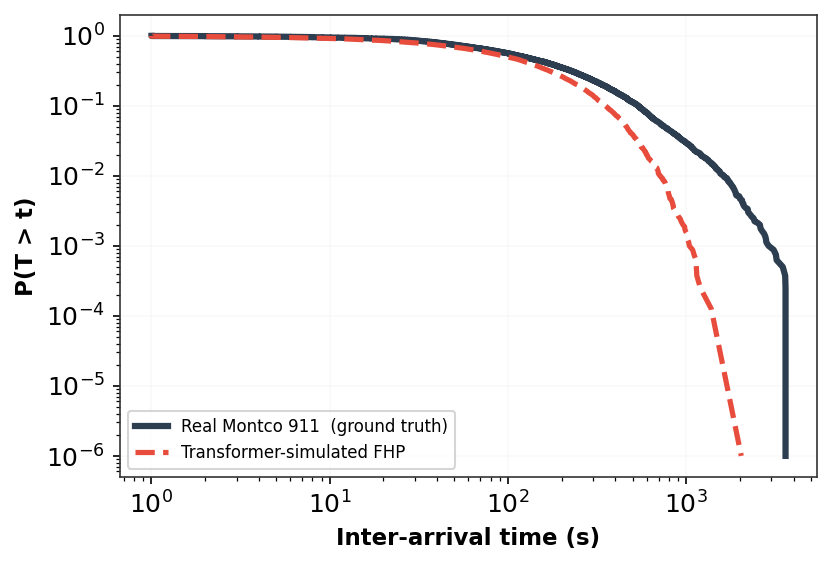}
    \label{fig:911_survival}
\end{subfigure}

\vspace{0.3cm}

\begin{subfigure}[t]{0.48\textwidth}
    \centering
    \includegraphics[width=\linewidth]{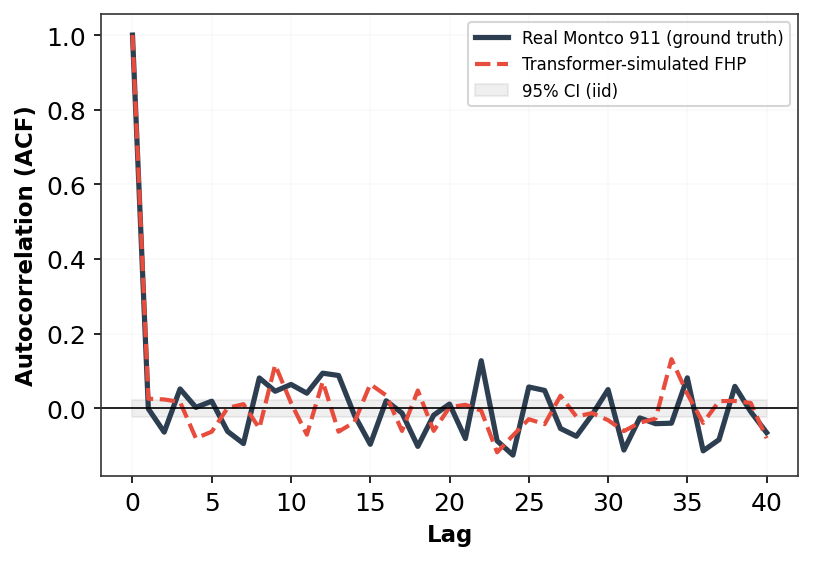}
    \label{fig:911_acf}
\end{subfigure}
\hfill
\begin{subfigure}[t]{0.48\textwidth}
    \centering
    \includegraphics[width=\linewidth]{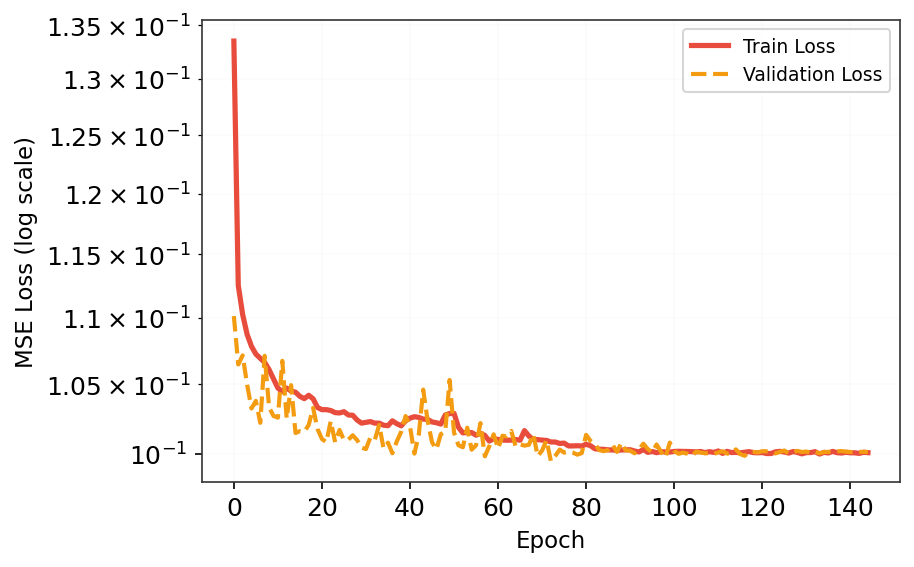}
    \label{fig:911_loss}
\end{subfigure}

\caption{Performance evaluation of the proposed Transformer-based FHP on the 911 dataset. (a) ECDF comparison between real and simulated inter-arrival times. (b) Survival probability comparison demonstrating the ability of the model to reproduce the tail behavior of the empirical distribution. (c) ACF comparison showing that the simulated sequence preserves the dependence structure observed in the real data. (d) Training and validation loss curves illustrating stable convergence and good generalization performance.}
\label{fig:911_results}

\end{figure}

The second data set, MontcoAlert 911, contains real emergency call records from Montgomery County, Pennsylvania, collected between $2015-2020$, including information on the time, location and category of emergency incidents. To evaluate the generalizability of the proposed Transformer-based FHP framework beyond financial event data, the model was trained using inter-arrival times extracted from the emergency call sequence. Figure~\ref{fig:911_results} presents the performance of the proposed approach in this data set. Figure~\ref{fig:911_results} top-left compares the ECDFs of the real and Transformer-simulated inter-arrival times. The two curves are in close agreement, with an ECDF distance of approximately $0.051$, indicating that the proposed model successfully reproduces the general distributional characteristics of the observed event arrivals. Figure~\ref{fig:911_results} top-right presents the survival probability comparison on a logarithmic scale. The simulated survival curve closely follows the empirical survival function over a broad range of inter-arrival times, demonstrating the capability of the model to capture both the bulk behavior and the tail properties of the emergency call process. To further assess the temporal dependence structure, Figure~\ref{fig:911_results} bottom-left compares the ACF of the real and Transformer-simulated inter-arrival sequences. The simulated sequence preserves the short-range dependence observed in the real data, and the estimated ACF exhibit a similar pattern across multiple lags. Finally, Figure~\ref{fig:911_results} bottom-right displays the training and validation loss curves of the Transformer-based FHP model over $150$ epochs. Both losses decrease steadily during training and converge to nearly identical values, indicating stable optimization and good generalization performance without noticeable overfitting. Overall, the results demonstrate that the proposed Transformer-based FHP framework effectively learns the underlying temporal dynamics of emergency call arrivals and generates synthetic event sequences that closely match the statistical distribution, tail behavior, and dependence structure of the real MontcoAlert 911 dataset.

\subsection{Effect of Window Size on Parameter Estimation and Dependence Structure}
\begin{figure}[htbp]
\centering

\begin{subfigure}[t]{0.32\textwidth}
    \centering
    \includegraphics[width=\linewidth]{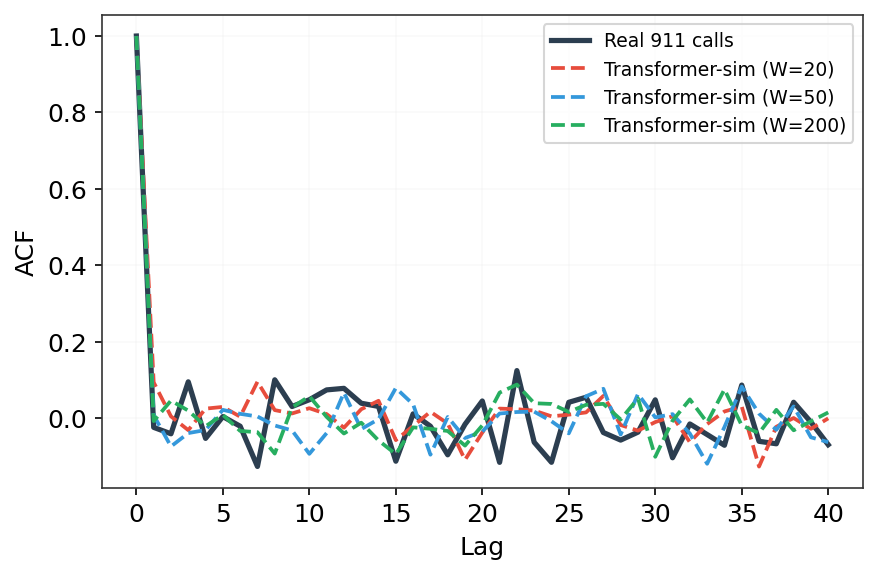}
\end{subfigure}
\hfill
\begin{subfigure}[t]{0.32\textwidth}
    \centering
    \includegraphics[width=\linewidth]{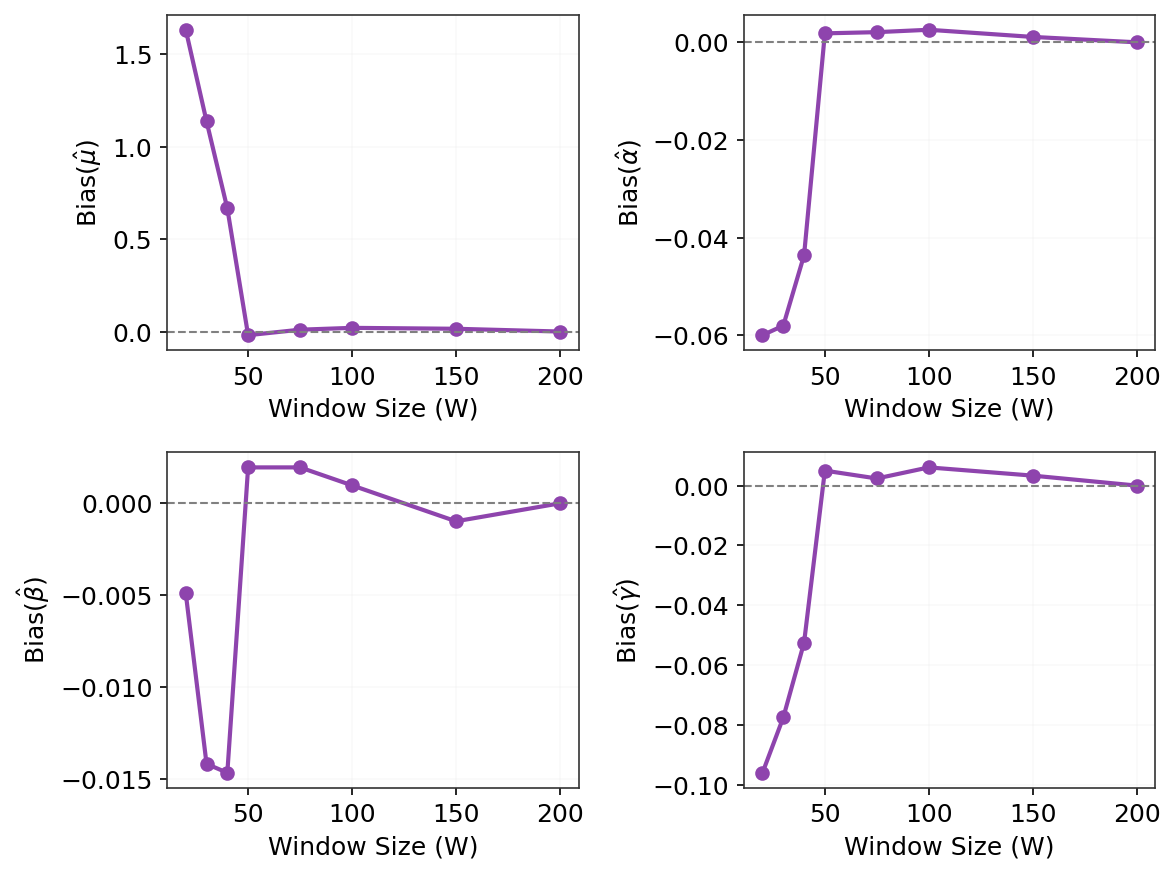}
\end{subfigure}
\hfill
\begin{subfigure}[t]{0.32\textwidth}
    \centering
    \includegraphics[width=\linewidth]{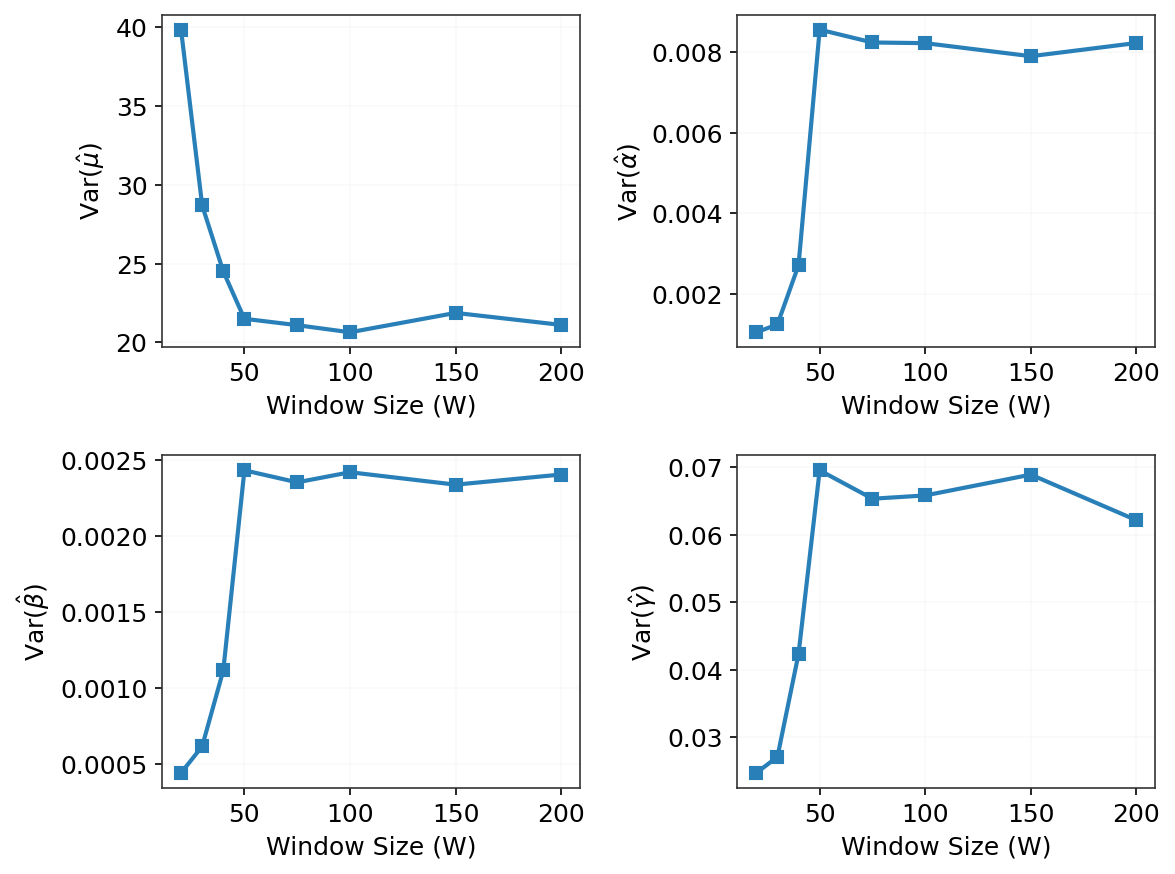}
    \label{fig:variance_911}
\end{subfigure}

\caption{Statistical analysis of the 911 event sequence: (a) autocorrelation function (ACF), (b) bias as a function of window size, and (c) variance as a function of window size.}
\label{fig:911_analysis}
\end{figure}

Furthermore, we investigate the effect of the sliding window size on parameter estimation and dependence preservation for the real AAPL and 911 event datasets.
Since the proposed Transformer-based estimation framework is based on fixed-length sliding windows, it is important to assess the impact of window size on estimation accuracy and dependence preservation. In self-exciting and long-memory processes such as the FHP, the amount of historical information contained within each window can significantly influence the bias and variance of the estimated parameters. Therefore, to evaluate the robustness of the proposed framework and justify the choice of window length, we performed a sensitivity analysis using window sizes
$$
W \in \{{20,35,50,75,100,150,200}\}
$$
for the AAPL and 911 datasets.
\begin{figure}[htbp]
\centering

\begin{subfigure}[t]{0.32\textwidth}
    \centering
    \includegraphics[width=\linewidth]{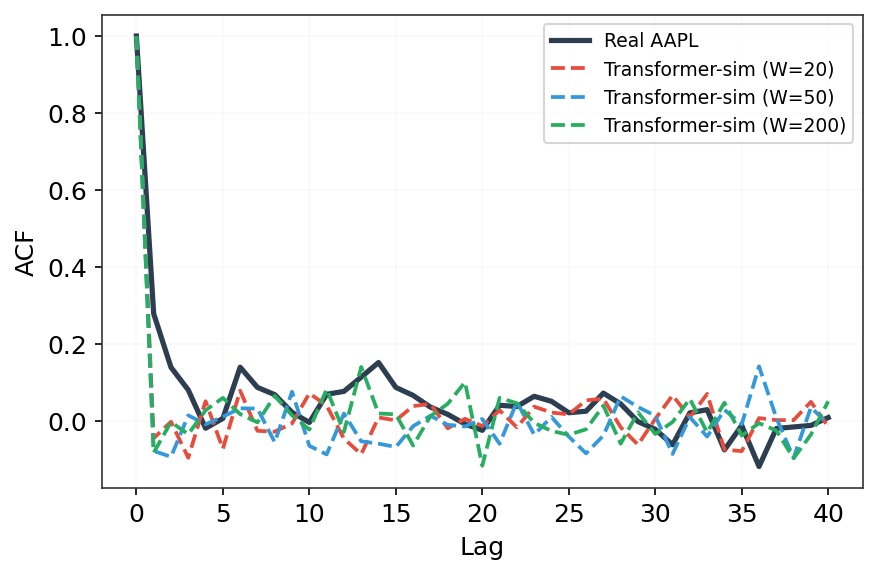}
\end{subfigure}
\hfill
\begin{subfigure}[t]{0.32\textwidth}
    \centering
    \includegraphics[width=\linewidth]{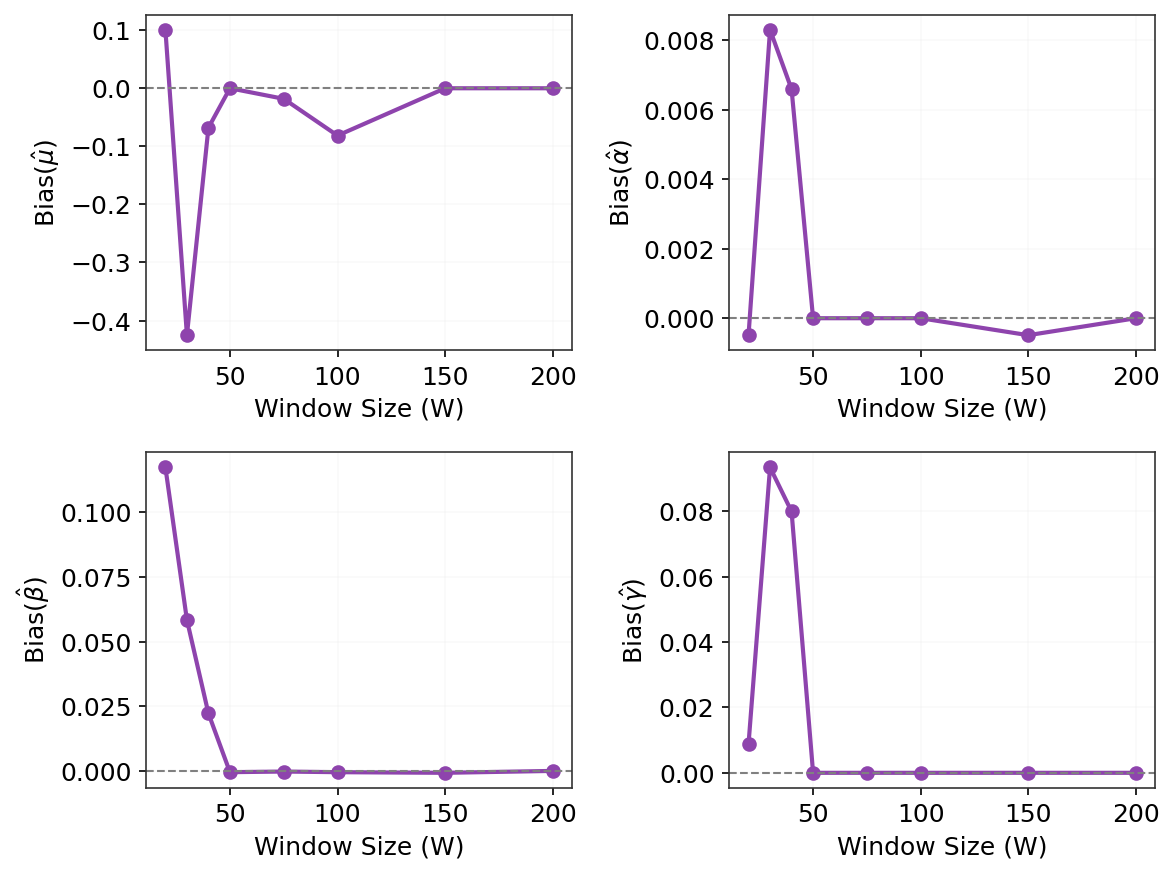}
\end{subfigure}
\hfill
\begin{subfigure}[t]{0.32\textwidth}
    \centering
    \includegraphics[width=\linewidth]{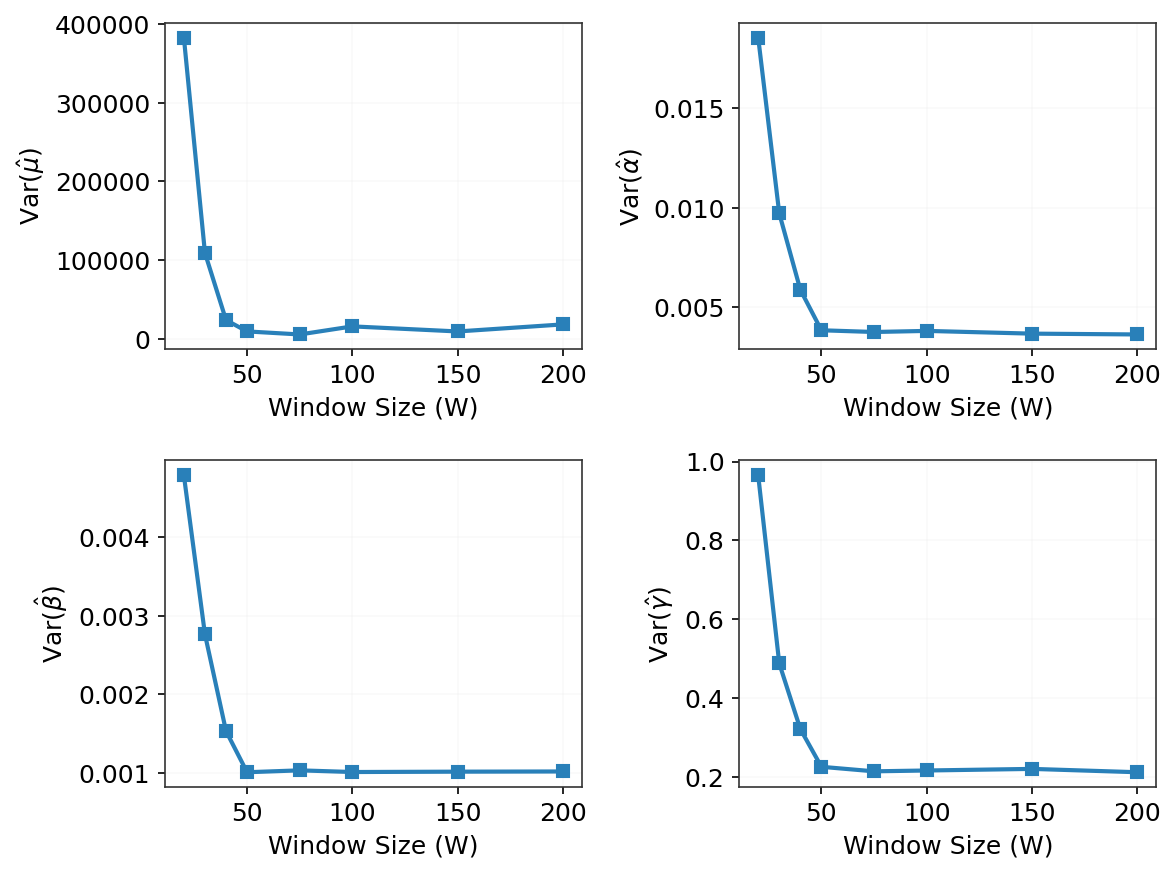}
\end{subfigure}

\caption{Statistical analysis of the AAPL event sequence: (a) autocorrelation function (ACF), (b) bias as a function of window size, and (c) variance as a function of window size.}
\label{fig:app_analysis}
\end{figure}
Figures~\ref{fig:app_analysis} and~\ref{fig:911_analysis} summarize the behavior of the estimation bias, variance, and ACF as the window size varies. The results show that smaller window sizes lead to larger estimation bias and higher variability, particularly for the baseline intensity parameter $\mu$, due to the limited temporal information available to the model. As the window size increases, both the bias and variance decrease substantially and stabilize for moderate and larger window lengths, indicating improved estimation reliability and reduced uncertainty. This behavior highlights the bias-variance trade-off associated with the choice of window size, where larger windows provide richer historical context and consequently more stable parameter estimates.
To further assess whether the temporal dependence structure is preserved, we compare the empirical ACF of the observed event sequences with the ACF obtained from Transformer-generated simulations across different window sizes. The simulated ACFs closely follow the empirical ACFs for both datasets and successfully reproduce the dominant dependence patterns across multiple lags. Although minor discrepancies are observed for smaller window sizes, larger windows provide a closer agreement with the empirical ACF, demonstrating that the proposed framework effectively captures the self-exciting and fractional dependence characteristics of the underlying process. Overall, the sensitivity analysis indicates that window sizes of approximately $W \geq 50$ provide a favorable balance between estimation accuracy, statistical stability, and preservation of the temporal dependence structure across different real-world event datasets.
\section{Conclusion}\label{section:6}
The proposed deep learning based \model framework using RNN (LSTM) and Transformer architectures for likelihood-free parameter estimation of FHP. The proposed models directly learn the relationship between inter-arrival time sequences and the underlying FHP parameters ($\mu,\gamma,\alpha,\beta$), thereby avoiding computationally intensive likelihood optimization. Experimental results on synthetic datasets demonstrated that both neural approaches outperform the classical MLE, with the Transformer consistently achieving the highest estimation accuracy due to its ability to capture long-range temporal dependencies through self-attention. The proposed framework was further validated on high-frequency AAPL NBBO transaction data and Montgomery County $911$ emergency call records, where the estimated parameters successfully reproduced the empirical distribution, tail behavior, and temporal dependence structure of the observed event sequences. In addition, the window-size sensitivity analysis showed that moderate to large window lengths provide more stable parameter estimates while preserving the dependence characteristics of the underlying process. Overall, the proposed Transformer-based framework provides an accurate and efficient alternative for parameter estimation in FHPs with long-memory dynamics. Future work will focus on extending the proposed approach to multivariate and marked FHPs, incorporating uncertainty quantification, and developing online learning frameworks for streaming event data.

\def\cprime{$'$}

\end{document}